\documentclass[11pt]{article}

\usepackage{amssymb}
\usepackage{amsmath,amsfonts}
\usepackage{booktabs}
\usepackage{array}
\usepackage[caption=false,font=normalsize,labelfont=rm,textfont=rm]{subfig}
\usepackage{textcomp}
\usepackage{stfloats}
\usepackage{url}
\usepackage{graphicx}
\usepackage[margin=1in]{geometry}
\usepackage{cite}
\graphicspath{{figures/}}
\DeclareGraphicsExtensions{.pdf,.png}
\usepackage{orcidlink}
\hypersetup{hidelinks}

\newcommand{\R}{\mathbb{R}}

\begin{document}

\title{Label-Decoupled Style Augmentation for Domain
Generalization in Multi-Label Remote Sensing Scene Classification}

\author{Alaa Almouradi$^{*}$
\and
        Erchan Aptoula
\thanks{This study was supported by the Scientific and Technological Research Council of Türkiye (TUBITAK) under the Grant Number
123R108.\\ Faculty of Engineering and Natural Sciences, Sabancı University, 34956 Istanbul, Türkiye (e-mail: alaaalmouradi@sabanciuniv.edu; erchan.aptoula@sabanciuniv.edu) \\ (\textit{Corresponding author: Alaa Almouradi.}) }}
\date{}
\maketitle
\begin{abstract}
Multi-label classification assigns several co-occurring labels to each aerial
scene, yet deployed models often encounter data distributions different from
their training. Feature-statistics augmentation such as MixStyle, EFDMix, and
correlated style uncertainty improves generalization at low cost but perturbs
channel statistics globally, treating each image as a single style; one class
can then contaminate the augmentation of another. Domain generalization
is understudied for multi-label remote sensing; no prior method or multi-source 
benchmark targets it. A label-decoupled
augmentation framework is therefore proposed, confining style perturbation to
label-specific regions. Per-label attention, obtained from a learnable module
or from gradient class-activation maps, yields per-label feature statistics;
these statistics are mixed with cross-domain samples that share present labels,
under independent per-label coefficients, and features are recomposed by
attention-weighted normalization. Three operators combined with two attention
sources produce six variants, evaluated on a leave-one-domain-out benchmark
from multi-label UCM, AID, and DFC15 over six shared labels. Averaged over
three splits and five seeds, the best variant attains 71.5\% mean average
precision, exceeding empirical risk minimization by 5.0 points and the
strongest global-statistics baseline by 1.3 points, with the largest gain on
the hardest transfer (up to 7.7 points). Ablations indicate that spatial
attention and refreshed localization maps are most influential. The framework
adds at most 0.35\% parameters, leaves inference unchanged, and appears to
offer a generic, inexpensive upgrade path for multi-label statistics-based domain
generalization. Code is available upon acceptance at
https://github.com/Alaa-Almouradi/Style-Augmentation-Upgrade.
\end{abstract}

% Attention, domain generalization, image classification, multi-label classification, remote sensing, style augmentation.

% ============================================================================
\section{Introduction}\label{sec:intro}
% ============================================================================

Aerial and satellite scenes rarely contain a single semantic category: a
typical urban tile simultaneously exhibits buildings, vehicles, vegetation,
and pavement. Multi-label classification, which assigns a set of co-occurring
labels to each image, is therefore a natural formulation for remote sensing
(RS) image understanding and supports applications such as land-use mapping,
urban monitoring, and content-based retrieval
\cite{cheng2017benchmark,chaudhuri2018ucmml,hua2019dfc15mlc}. Deep
convolutional networks have become the dominant approach to this task
\cite{zeggada2017uav,sumbul2020multiattention,hua2020aidml}, building on
backbone architectures pre-trained on large-scale natural-image collections
\cite{he2016resnet,deng2009imagenet}.

These models, however, are typically trained and evaluated under the
assumption of independent and identically distributed data. In practice, a
classifier trained on imagery from one sensor, region, season, or ground sampling
distance is frequently applied to imagery acquired under different conditions,
and the resulting distribution shift degrades performance considerably
\cite{tuia2016da}. Domain adaptation techniques mitigate the shift by
exploiting target-domain data during training
\cite{ganin2016dann,tuia2016da}, but such data is often unavailable before
deployment. Domain generalization (DG) addresses the harder setting in which
a model is trained on several source domains and must perform well on an
unseen target domain
\cite{blanchard2011dg,zhou2023dgsurvey,wang2023dgsurvey}. Among the many DG
families (invariant representation learning \cite{muandet2013dica},
adversarial data augmentation \cite{volpi2018adversarial}, self-supervision
\cite{carlucci2019jigsaw}, frequency-domain augmentation \cite{xu2021fact},
style-bias reduction \cite{nam2021sagnet}, and flat-minima optimization
\cite{cha2021swad}) feature-statistics augmentation has proven particularly
attractive; methods such as MixStyle \cite{zhou2021mixstyle}, DSU
\cite{li2022dsu}, EFDMix \cite{zhang2022efdm}, and correlated style
uncertainty (CSU) \cite{zhang2024csu} perturb instance-level feature
statistics during training, are inserted into existing backbones with a few
lines of code, and add essentially no inference cost.

However, these augmentation methods share an implicit assumption that limits
their suitability for multi-label imagery: they operate on \emph{global}
per-channel statistics, treating each image as carrying a single homogeneous
style. This assumption is reasonable for object-centric, single-label
benchmarks, but a multi-label RS tile is a mosaic of regions whose local
statistics differ substantially; water bodies, rooftops, and vegetation
produce markedly different channel responses. Mixing global statistics
therefore entangles label-specific appearance. Consider a coastal tile in which
\emph{ship}, \emph{water}, and \emph{pavement} co-occur: a global statistics
swap with a partner image can repaint the dark, low-variance signature of open
water onto the hulls of the ships and the quay, and because a single mixing
coefficient is shared by the whole image, the perturbation applied to the rare
\emph{ship} class is locked to that of the dominant \emph{pavement}, so
augmenting one class inevitably distorts the others. More generally, the style
of a water region in one image may be transferred onto the building regions of
another, and the coupled coefficient prevents the perturbation strength of
co-occurring classes from being controlled independently. In addition, DG itself remains
largely unexplored for multi-label RS classification. Recent surveys of DG in
remote sensing \cite{liao2026rsdgsurvey} catalogue methods for single-label
scene classification, object detection, change detection, and semantic
segmentation, but none for multi-label scene classification, and no established
multi-source evaluation benchmark exists for the latter; most multi-label RS
studies evaluate within a single dataset
\cite{zeggada2017uav,koda2018ssvm,hua2020aidml}, and cross-dataset
robustness is rarely quantified in the DG setting.
The present study is the first to address DG for multi-label RS scene
classification and to provide a multi-source multi-label benchmark for it.

To address these limitations, this paper proposes a \emph{label-decoupled}
(LD) style augmentation framework that confines statistics perturbation to
label-specific spatial regions. Per-label attention maps are produced either by
a lightweight learnable label-localization attention module (LLAM) or by a
periodically refreshed, parameter-free Grad-CAM \cite{selvaraju2017gradcam}
bank. These per-label attention maps decompose each feature map into per-label
statistics. These statistics
are then mixed exclusively between cross-domain samples that share the
corresponding label, each label receiving an independent mixing coefficient,
and the perturbed components are recomposed through attention-weighted
normalization. The hypothesis is that confining style transfer to matching
semantic regions yields more realistic style diversity and, hence, better
generalization to unseen domains; this hypothesis is examined through
controlled comparisons, ablations, and an analysis of the learned style
manifolds. The main contributions are as follows:
\begin{enumerate}
\item A label-decoupled augmentation framework comprising attention-driven
per-label statistic decomposition, cross-domain label-matched partner
selection, per-label mixing coefficients, and attention-weighted
recomposition. The framework is instantiated on three established operators
(MixStyle \cite{zhou2021mixstyle}, EFDMix \cite{zhang2022efdm}, and CSU
\cite{zhang2024csu}) with two attention sources, yielding six variants. \textbf{This is the first domain generalization method designed
specifically for \textit{multi-label} RS scene classification.}
\item Supporting components that make the decomposition practical: a per-label
cross-domain style memory bank, a global-to-per-label warm-up schedule, an
attention diversity regularizer, and a label-matched domain-balanced sampler.
\item A leave-one-domain-out multi-label DG benchmark assembled from three
public multi-label RS datasets (UCM, AID, and DFC15
\cite{yang2010ucm,chaudhuri2018ucmml,xia2017aid,hua2020aidml,dfc15,hua2019dfc15mlc})
over six shared labels, which is the first
multi-source multi-label domain generalization benchmark in remote sensing.
\item An extensive empirical study (three leave-one-domain-out splits, five seeds, eleven
methods, component ablations, hyperparameter sensitivity, manifold
visualization, and complexity analysis) indicating consistent improvements
over the corresponding global-statistics methods at negligible cost.
\end{enumerate}

The remainder of this paper is organized as follows.
Section~\ref{sec:related} reviews related work. Section~\ref{sec:method}
presents the proposed framework. Section~\ref{sec:setup} describes the
benchmark and experimental setup, and Section~\ref{sec:results} reports
and discusses the results. Section~\ref{sec:conclusion} concludes the paper.

% ============================================================================
\section{Related work}\label{sec:related}
% ============================================================================

\subsection{Multi-label remote sensing image classification}

Multi-label learning generalizes single-label classification by allowing each
sample to carry a set of labels
\cite{tsoumakas2007multilabel,zhang2014review}. In computer vision, deep
multi-label models have evolved from independent binary classifiers toward
architectures that exploit label dependencies and spatial structure,
e.g.~recurrent label decoding \cite{wang2016cnnrnn}, spatial regularization
\cite{zhu2017srn}, and label graphs \cite{chen2019mlgcn}. In RS, early deep
approaches adapted convolutional networks to aerial multi-labeling
\cite{zeggada2017uav}, structured support vector machines incorporated
spatial label correlations \cite{koda2018ssvm}, and data augmentation was
shown to benefit multi-label land-cover categorization
\cite{stivaktakis2019augmentation}. In parallel, attention has played an increasingly
central role: class-wise attention combined with recurrent modeling
\cite{hua2019dfc15mlc}, multi-attention feature extraction
\cite{sumbul2020multiattention}, and attention-driven unmanned aerial vehicle (UAV) image
multi-labeling \cite{alshehri2019attention} all localize label-relevant
evidence, while relation networks \cite{hua2020aidml} and graph
convolutional models \cite{khan2019gcn} encode inter-label dependencies.
Benchmarks such as multi-label UCM \cite{yang2010ucm,chaudhuri2018ucmml},
multi-label AID \cite{xia2017aid,hua2020aidml}, and the DFC15 multi-label
set \cite{dfc15,hua2019dfc15mlc} have standardized within-dataset
evaluation. Nevertheless, these studies almost exclusively train and test on
the same distribution; the behavior of multi-label RS classifiers under
domain shift has received little systematic attention in the DG setting, which motivates 
the benchmark used in this work.

\subsection{Domain generalization}

DG aims to learn, from one or several source domains, a predictor that transfers
to unseen target domains without access to target data during training
\cite{blanchard2011dg}. Comprehensive surveys categorize the field into
representation-based, data-augmentation-based, and learning-strategy-based
approaches \cite{zhou2023dgsurvey,wang2023dgsurvey}. Representative directions
include domain-invariant component analysis \cite{muandet2013dica}, adversarial
feature alignment originally developed for adaptation \cite{ganin2016dann},
adversarial data augmentation \cite{volpi2018adversarial}, self-supervised
auxiliary tasks \cite{carlucci2019jigsaw}, Fourier-based amplitude mixing
\cite{xu2021fact}, style-bias reduction \cite{nam2021sagnet}, and stochastic
weight averaging densely (SWAD), which seeks flat minima of the loss landscape
\cite{cha2021swad}.

Most of these methods are single-label by construction. \emph{Multi-label} DG,
in which each image carries several co-occurring labels, has received far less
attention, and the few existing approaches are dominated by vision-language
models, which operate on a different set of constraints. CLIPood adapts a pretrained contrastive language--image model to
out-of-distribution data through a margin-based objective and a Beta-weighted
parameter average \cite{shu2023clipood}, and Mixup-CLIPood extends this idea with 
a class-aware mix-up loss over larger
language--image backbones \cite{qiao2024mixupclipood}. Such methods inherit the
broad coverage of web-scale pretraining and the multi-modal advantage, but they depend on heavy foundation
backbones, are designed for generic object imagery rather than overhead scenes,
and still characterize style at the image level rather than per label.

Within remote sensing, distribution shift has historically been tackled by
domain adaptation, which assumes access to (typically unlabeled) target data
\cite{tuia2016da}. RS DG has gained traction only recently; a dedicated survey
\cite{liao2026rsdgsurvey} organizes RS DG methods into distribution alignment,
data augmentation, and representation learning across scene classification,
object detection, change detection, and semantic segmentation. Style and
feature augmentation is, in fact, a recurring tool in this literature: SODA
randomizes styles with adaptive instance normalization \cite{zhao2024soda},
FOSMix mixes style in the frequency domain for RS segmentation
\cite{iizuka2023fosmix}, and a style-content separation network disentangles
style from content for cross-scene generalization \cite{zhu2023scsn}. Other
lines align spectral bands across countries \cite{durakli2024badg}, collaborate
across multiple sources \cite{han2024mscdg}, generalize object detectors
\cite{durakli2023roofdgod}, guide invariance with language
\cite{zhang2023ldgnet}, or fine-tune geospatial vision foundation models
\cite{gong2025crossearth}. Crucially, every one of these methods operates on
single-label scene classification, detection, change detection, segmentation,
or pixel-wise hyperspectral classification; the multi-label
scene-classification setting, in which several classes with distinct local
styles share a single image, is addressed by none of them, and the benchmarks
they adopt are correspondingly single-label \cite{liao2026rsdgsurvey}.

These observations expose a specific gap. Global feature-statistics augmentation
is cheap and effective but is single-style per image by construction; multi-label
DG techniques exist but are vision-language and not tailored to overhead
imagery; and RS DG, despite using style augmentation extensively, has not
addressed multi-label scene classification. The framework proposed here targets
precisely this intersection by decoupling style statistics per label, which is the first such treatment for multi-label RS DG.

\subsection{Feature-statistics style augmentation}

Instance-level feature statistics encode image style: adaptive instance
normalization (AdaIN) transfers style by replacing the per-channel mean and
standard deviation of one image's features with those of another
\cite{huang2017adain}. Building on this observation, MixStyle interpolates
the feature statistics of random sample pairs during training
\cite{zhou2021mixstyle}; DSU models statistics as Gaussian random variables
and resamples them \cite{li2022dsu}; EFDMix matches higher-order statistics
exactly through sorted-value interpolation \cite{zhang2022efdm}; and CSU
perturbs statistics with channel-correlated uncertainty \cite{zhang2024csu}.
These methods are attractive for their simplicity, negligible cost, and
strong results on single-label DG benchmarks. However, all of them compute
statistics by pooling over the \emph{entire} spatial extent of the feature
map, which presumes one dominant style per image, and all of them draw a
\emph{single} mixing coefficient per image, which couples the perturbation
strength of every class in the scene. Both design choices conflict with the
structure of multi-label aerial imagery, in which several classes with
distinct local statistics co-occur. The framework proposed in this paper removes both
restrictions by decomposing statistics per label with spatial
attention, obtained from a learned module or from class activation maps
\cite{zhou2016cam,selvaraju2017gradcam}, and by mixing each label's
statistics independently.

% ============================================================================
\section{Proposed approach}\label{sec:method}
% ============================================================================

\subsection{Overview and problem formulation}\label{sec:overview}

Let $\mathcal{D}_1,\dots,\mathcal{D}_S$ denote $S$ source domains that share a
label vocabulary of $L$ classes. Each training sample is a triplet
$(\mathbf{x}_i,\mathbf{y}_i,d_i)$, where $\mathbf{x}_i\in\R^{3\times H_0\times
W_0}$ is an image, $\mathbf{y}_i\in\{0,1\}^{L}$ is its multi-label annotation,
and $d_i\in\{1,\dots,S\}$ is its domain index. The goal is to learn a
multi-label classifier $g:\R^{3\times H_0\times W_0}\to[0,1]^{L}$ that
performs well on an unseen target domain.

\begin{figure*}[t] 
\begin{center}
\includegraphics[width=\textwidth]{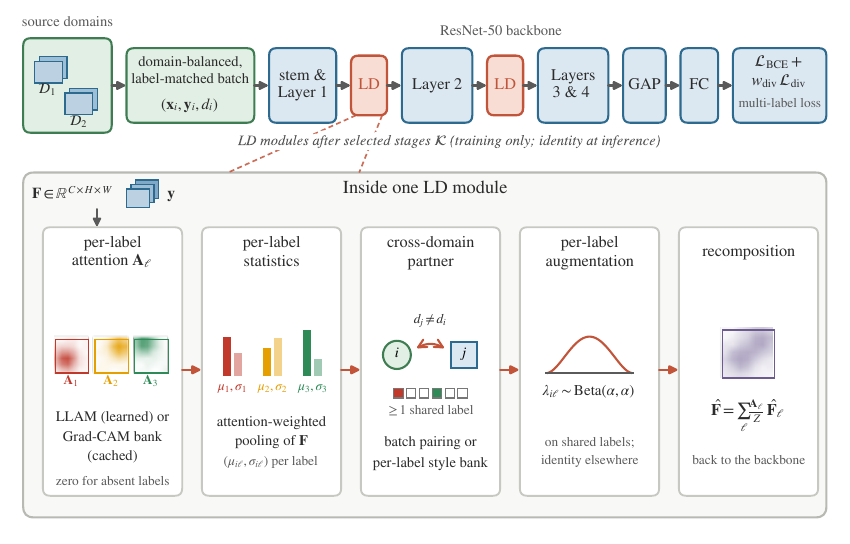}
\end{center}
\caption{Overview of the proposed LD style augmentation
framework. Top: training pipeline with LD modules inserted after selected
backbone layers $\mathcal{K}$; the modules are active only during training
and reduce to the identity at inference. Bottom: inside one LD module,
per-label attention (learned LLAM or cached Grad-CAM) yields per-label
statistics, which are mixed only with cross-domain partners sharing the
corresponding label (through batch pairing or a per-label style bank),
perturbed with independent per-label coefficients, and recomposed by
attention-weighted normalization.}\label{fig:pipeline}
\end{figure*}

Fig.~\ref{fig:pipeline} summarizes the proposed framework. A standard
backbone is augmented with LD modules inserted after each layer in a set
$\mathcal{K}$ of selected layers. During training, each module
(i)~estimates per-label spatial attention, (ii)~pools per-label feature
statistics, (iii)~selects a cross-domain partner that shares at least one
label (or draws stored statistics from a per-label style bank),
(iv)~perturbs each label's statistics independently with one of three
operators, and (v)~recomposes the feature map through attention-weighted
normalization. The modules are stochastic and active only during training;
at inference they reduce to the identity, so the deployed network is exactly
the original backbone.

\subsection{Per-label decomposition and recomposition}\label{sec:decomp}

Consider the feature map $\mathbf{F}_i\in\R^{C\times H\times W}$ of sample
$i$ at an insertion point, and let
$\mathbf{A}_i\in[0,1]^{L\times H\times W}$ collect per-label attention maps
(Section~\ref{sec:attention}), with $\mathbf{A}_{i\ell}=\mathbf{0}$ for
absent labels ($y_{i\ell}=0$). Normalized spatial weights are obtained as
\begin{equation}\label{eq:attnorm}
\bar{A}_{i\ell hw} \;=\;
\frac{A_{i\ell hw}}{\sum_{h'=1}^{H}\sum_{w'=1}^{W} A_{i\ell h'w'}
+ \varepsilon},
\end{equation}
where $\varepsilon$ is a small constant for numerical stability. The
attention-weighted first and second moments of each present label $\ell$ are
\begin{equation}\label{eq:stats}
\begin{aligned}
\mu_{i\ell c} &= \sum_{h,w} \bar{A}_{i\ell hw}\, F_{ichw},\\
\sigma_{i\ell c} &= \Big(\sum_{h,w} \bar{A}_{i\ell hw} F_{ichw}^{2}
 - \mu_{i\ell c}^{2} + \varepsilon\Big)^{\!1/2}\!,
\end{aligned}
\end{equation}
yielding per-label style vectors
$\boldsymbol{\mu}_{i\ell},\boldsymbol{\sigma}_{i\ell}\in\R^{C}$. Given
perturbed statistics
$(\hat{\boldsymbol{\mu}}_{i\ell},\hat{\boldsymbol{\sigma}}_{i\ell})$ produced
by one of the operators in Section~\ref{sec:operators}, the augmented feature
map is recomposed as a convex spatial combination of per-label stylized
components,
\begin{equation}\label{eq:recomp}
\hat{\mathbf{F}}_{i} = \sum_{\ell=1}^{L}
\frac{\mathbf{A}_{i\ell}}{\mathbf{Z}_i}\odot
\Big(
\hat{\boldsymbol{\sigma}}_{i\ell}
\frac{\mathbf{F}_{i}-\boldsymbol{\mu}_{i\ell}}{\boldsymbol{\sigma}_{i\ell}}
+\hat{\boldsymbol{\mu}}_{i\ell}
\Big),
\quad
\mathbf{Z}_i=\sum_{\ell=1}^{L}\mathbf{A}_{i\ell}+\varepsilon,
\end{equation}
where $\odot$ denotes element-wise multiplication with broadcasting of the
channel-wise statistics over spatial locations.
Equation~\eqref{eq:recomp} performs, per label, the
normalization--renormalization of AdaIN \cite{huang2017adain}, but blends
the results spatially according to the attention, so that each region is
restyled predominantly by the statistics of the labels that occupy it.
Locations not covered by any present label are dominated by the residual
normalization term and remain close to the original features; labels whose
statistics are left unperturbed (Section~\ref{sec:pairing}) contribute
identity transformations. Each per-label branch is a channel-wise affine map of
the standardized features, i.e.~it preserves the spatial geometry of
$\mathbf{F}_i$ and only rescales and shifts channels; the convex,
attention-weighted combination of such branches therefore yields a
well-defined feature tensor that stays on the same manifold as a single AdaIN
restyling, rather than an arbitrary distortion that the network could simply
learn to ignore.

\subsection{Per-label attention}\label{sec:attention}

Two interchangeable attention sources are considered.

\paragraph{Learned attention using LLAM}
The label-localization attention module is a lightweight head consisting of
two $1\times1$ convolutions with an intermediate rectified linear unit,
mapping $\mathbf{F}_i$ to logits $\mathbf{z}_i\in\R^{L\times H\times W}$,
followed by a temperature-sharpened spatial softmax and masking by the
ground-truth label vector:
\begin{equation}\label{eq:llam}
A_{i\ell hw} \;=\; y_{i\ell}\,
\frac{\exp\!\big(\tau\, z_{i\ell hw}\big)}
{\sum_{h',w'}\exp\!\big(\tau\, z_{i\ell h'w'}\big)},
\end{equation}
where $\tau\geq 1$ controls map sharpness. Because co-occurring labels should
attend to different regions, a diversity regularizer penalizes the mean
pairwise cosine similarity between the attention maps of labels that are
jointly present:
\begin{equation}\label{eq:div}
\mathcal{L}_{\mathrm{div}}
=\frac{1}{B}\sum_{i=1}^{B}
\frac{1}{|\mathcal{P}_i|}
\sum_{(\ell,m)\in\mathcal{P}_i}
\frac{\langle \mathbf{a}_{i\ell},\mathbf{a}_{im}\rangle}
{\lVert\mathbf{a}_{i\ell}\rVert_2\,\lVert\mathbf{a}_{im}\rVert_2},
\end{equation}
where $B$ is the batch size, $\mathbf{a}_{i\ell}\in\R^{HW}$ is the flattened
map of label $\ell$, and
$\mathcal{P}_i=\{(\ell,m):\ell<m,\ y_{i\ell}=y_{im}=1\}$. In the
configuration of Section~\ref{sec:results}, the module adds
$8.3\times10^{4}$ parameters ($+0.35\%$).

\paragraph{Cached Grad-CAM bank (GC)}
As a parameter-free alternative, class activation maps
\cite{zhou2016cam,selvaraju2017gradcam} of the network's own predictions are
used. For label $\ell$ of sample $i$, the map is
\begin{equation}\label{eq:gradcam}
\begin{split}
\mathbf{A}^{\mathrm{cam}}_{i\ell}
&=\mathrm{ReLU}\!\Big(\sum_{c=1}^{C} w_{i\ell c}\,\mathbf{F}^{(4)}_{ic}\Big),\\
w_{i\ell c}
&=\frac{1}{HW}\sum_{h,w}
\frac{\partial s_{i\ell}}{\partial F^{(4)}_{ichw}},
\end{split}
\end{equation}
where $\mathbf{F}^{(4)}_i$ is the final-stage feature map and $s_{i\ell}$
the pre-sigmoid logit of label $\ell$. Computing \eqref{eq:gradcam} online at
every step would be costly and, early in training, unreliable. Instead, a
bank of maps is built once the warm-up ends and is refreshed every $R$ epochs
by a single sweep over the training set; an entry is stored only if label
$\ell$ is present \emph{and} currently predicted correctly, is kept in half
precision on the host, and is bilinearly resized and renormalized to the
resolution of each insertion point when used. Missing entries
(e.g.~not-yet-correct labels) fall back to the global pathway for that label.
This correctness-gated caching is also what guards the decomposition against the
well-known noisiness of class activation maps and their sensitivity to
distribution shift: a map is admitted only where the network already localizes
the label well on a \emph{source} image, maps are refreshed as the model
improves, and any label without a reliable entry reverts to the global operator,
so a poorly localized map degrades gracefully toward the global baseline rather
than injecting cross-class contamination. Because the per-label components are
finally recombined by attention-weighted normalization
\eqref{eq:recomp}, residual localization error is further attenuated by the
spatial blending. The learned LLAM source offers a complementary safeguard, as
its maps are optimized end-to-end together with the mixing.
The two attention sources trade off a small parameter and compute overhead
with a learned, task-adaptive map (LLAM) against zero parameters with a
periodic refresh cost (GC).

\subsection{Cross-domain label-matched pairing and style bank}\label{sec:pairing}

Global-statistics methods mix arbitrary sample pairs. Here, mixing partners
are constrained to differ in domain and to share semantic content: sample $i$
is paired with a partner $j=\pi(i)$ drawn from
\begin{equation}\label{eq:partner}
\mathcal{P}(i)=\big\{\,j:\ d_j\neq d_i\ \text{and}\
\mathbf{y}_i^{\top}\mathbf{y}_j\geq 1 \,\big\},
\end{equation}
computed within each mini-batch by a greedy matching over a domain-balanced
sampler that explicitly co-locates label-compatible samples from different
domains. The per-label shared mask and the effective mixing coefficient are
\begin{equation}\label{eq:mask}
m_{i\ell}=y_{i\ell}\,y_{j\ell},
\qquad
\tilde{\lambda}_{i\ell}= m_{i\ell}\,\lambda_{i\ell} + (1-m_{i\ell}),
\end{equation}
where $\lambda_{i\ell}\sim\mathrm{Beta}(\alpha,\alpha)$, so that labels not
shared with the partner receive
$\tilde{\lambda}_{i\ell}=1$, i.e.~the identity, and every shared label draws
an \emph{independent} coefficient. For the interpolation operator, partners
may alternatively be drawn from a per-label first-in-first-out style bank of
size $M$ that stores recent
$(\boldsymbol{\mu}_{\cdot\ell},\boldsymbol{\sigma}_{\cdot\ell},d)$ triplets
and is sampled under the same cross-domain constraint, decoupling the
diversity of styles from the mini-batch composition; when a label has no
cross-domain entry yet, batch pairing is used as a fallback.

\subsection{Per-label augmentation operators}\label{sec:operators}

\begin{figure*}[t]
\begin{center}
\includegraphics[width=\textwidth]{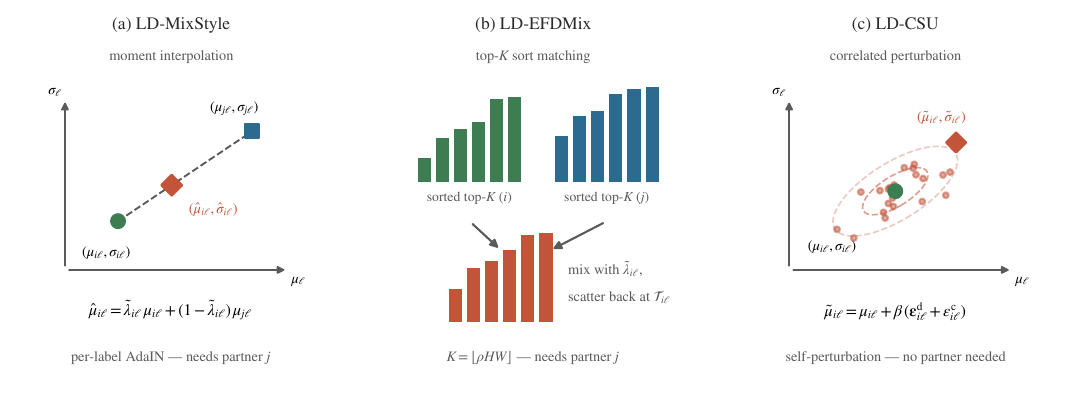}
\end{center}
\caption{The three per-label augmentation operators explained in subsection \ref{sec:operators}. (a)~LD-MixStyle
interpolates the per-label moments of sample $i$ toward those of a
cross-domain partner $j$ sharing label $\ell$. (b)~LD-EFDMix sorts the
feature values at the top-$K$ attention locations of both samples and mixes
them in rank order. (c)~LD-CSU perturbs the per-label moments with
channel-correlated Gaussian noise and requires no partner.}
\label{fig:operators}
\end{figure*}

Three operators, illustrated in Fig.~\ref{fig:operators}, instantiate the
framework; each is applied per present label with the coefficient
$\tilde{\lambda}_{i\ell}$ of \eqref{eq:mask}.

\paragraph{LD-MixStyle}
Following the interpolation principle of Mix\-Style \cite{zhou2021mixstyle},
the perturbed statistics are convex combinations of the per-label statistics
of the two samples:
\begin{equation}\label{eq:mixstyle}
\begin{aligned}    
\hat{\boldsymbol{\mu}}_{i\ell}
=\tilde{\lambda}_{i\ell}\boldsymbol{\mu}_{i\ell}
+(1-\tilde{\lambda}_{i\ell})\boldsymbol{\mu}_{j\ell},\\
\quad
\hat{\boldsymbol{\sigma}}_{i\ell}
=\tilde{\lambda}_{i\ell}\boldsymbol{\sigma}_{i\ell}
+(1-\tilde{\lambda}_{i\ell})\boldsymbol{\sigma}_{j\ell},
\end{aligned}
\end{equation}
These values are inserted into the recomposition \eqref{eq:recomp}. The partner statistics
come from the paired sample or from the style bank.

\paragraph{LD-EFDMix}
Moment interpolation captures only first- and second-order statistics; exact
feature distribution matching \cite{zhang2022efdm} instead operates on
sorted feature values. Let $\mathcal{T}_{i\ell}$ contain the
$K=\lfloor\rho HW\rfloor$ spatial locations with the largest attention
$A_{i\ell hw}$, and let
$\mathbf{v}^{c}_{i\ell}=\mathrm{sort}\big(\{F_{ichw}:(h,w)\in
\mathcal{T}_{i\ell}\}\big)$ denote the sorted values of channel $c$ at those
locations (analogously for the partner). The mixed sequence
\begin{equation}\label{eq:efdmix}
\hat{\mathbf{v}}^{c}_{i\ell}
=\tilde{\lambda}_{i\ell}\,\mathbf{v}^{c}_{i\ell}
+(1-\tilde{\lambda}_{i\ell})\,\mathbf{v}^{c}_{j\ell}
\end{equation}
is scattered back to the original locations of sample $i$ in rank order, the
partner term being treated as a constant (gradient-stopped) as in
\cite{zhang2022efdm}; the resulting per-label value changes are then blended
spatially with the attention weights, analogously to \eqref{eq:recomp}.
Restricting the matching to the top-$K$ locations of \emph{both} attention
maps confines the exchanged distribution to label-relevant evidence.

\paragraph{LD-CSU}
Inspired by correlated style uncertainty \cite{zhang2024csu}, this operator
perturbs each label's statistics without any partner. With
$\boldsymbol{\xi}\sim\mathcal{N}(\mathbf{0},\mathbf{I}_C)$ and
$\eta\sim\mathcal{N}(0,1)$,
\begin{equation}\label{eq:csu}
\begin{split}
\boldsymbol{\varepsilon}^{\mathrm{d}}_{i\ell}
&=\boldsymbol{\sigma}_{i\ell}\odot\boldsymbol{\xi},
\qquad
\varepsilon^{\mathrm{c}}_{i\ell}
=\bar{\sigma}_{i\ell}\,\eta,\\
\tilde{\boldsymbol{\mu}}_{i\ell}
&=\boldsymbol{\mu}_{i\ell}
+\beta\big(\boldsymbol{\varepsilon}^{\mathrm{d}}_{i\ell}
+\varepsilon^{\mathrm{c}}_{i\ell}\mathbf{1}\big),\\
\tilde{\boldsymbol{\sigma}}_{i\ell}
&=\max\!\big(\boldsymbol{\sigma}_{i\ell}
+\beta\,\lvert\boldsymbol{\varepsilon}^{\mathrm{d}}_{i\ell}\rvert,
\varepsilon\big),
\end{split}
\end{equation}
where $\bar{\sigma}_{i\ell}$ is the channel-mean standard deviation,
$\mathbf{1}\in\R^{C}$ is a vector of ones, and $\beta$ is the perturbation
strength. The scalar term $\varepsilon^{\mathrm{c}}_{i\ell}$ displaces all
channels jointly, injecting the channel-correlated component advocated in
\cite{zhang2024csu} at the per-label level.

Combining the three operators with the two attention sources of
Section~\ref{sec:attention} yields six variants, denoted LD-MixStyle,
LD-EFDMix, LD-CSU (learned attention) and LD-MixStyle-GC, LD-EFDMix-GC,
LD-CSU-GC (Grad-CAM bank).

\subsection{Training procedure}\label{sec:training}

Each LD module fires stochastically with probability $p$ per batch, as in
\cite{zhou2021mixstyle}. Reliable per-label attention is unavailable at the
start of training: LLAM logits are untrained and network predictions, on
which the Grad-CAM bank conditions, are still poor. Training therefore
begins in a \emph{global} mode in which the modules behave exactly like their
global-statistics counterparts. For LLAM variants, the per-label pathway is
blended in linearly,
\begin{equation}\label{eq:warmup}
\begin{split}
\mathbf{F}^{\mathrm{out}}_{i}
&= w_e\,\hat{\mathbf{F}}^{\mathrm{LD}}_{i}
+(1-w_e)\,\hat{\mathbf{F}}^{\mathrm{glob}}_{i},\\
w_e
&=\min\!\big(\max\!\big((e-W)/W,\,0\big),\,1\big),
\end{split}
\end{equation}
where $e$ is the epoch index and $W$ is the warm-up length; GC variants
switch to the per-label pathway as soon as the first bank has been built
after epoch $W$. The total objective is
\begin{equation}\label{eq:loss}
\mathcal{L}
=\mathcal{L}_{\mathrm{BCE}}
+ w_{\mathrm{div}}\,\mathcal{L}_{\mathrm{div}},
\end{equation}
where $\mathcal{L}_{\mathrm{BCE}}$ is the mean binary cross-entropy over
labels and $w_{\mathrm{div}}=0$ for GC variants, whose attention is not
learned. At inference all modules are inactive, so the computational graph
is identical to the plain backbone.

% ============================================================================
\section{Experimental setup}\label{sec:setup}
% ============================================================================

\subsection{Datasets and splits}\label{sec:datasets}

\begin{figure*}[t]
\begin{center}
\includegraphics[width=\textwidth]{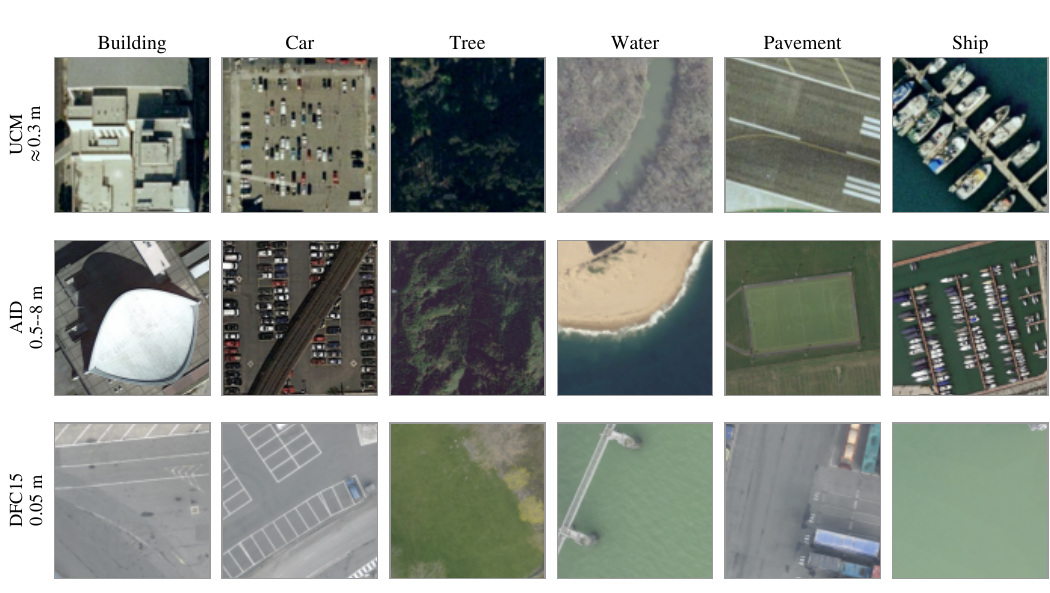}
\end{center}
\caption{Class samples from each domain in the benchmark, where the domain shift can be seen through the variations between images of the same class from different domains.}
\label{fig:domain_samples}
\end{figure*}

The benchmark is assembled from three public multi-label aerial datasets
whose label vocabularies overlap in six classes: \emph{building}, \emph{car},
\emph{tree}, \emph{water}, \emph{pavement}, and \emph{ship}.
UCM \cite{yang2010ucm} with the multi-label annotations of
\cite{chaudhuri2018ucmml} contains 2100 images of $256\times256$ pixels at
approximately $0.3$~m ground sampling distance (GSD). AID \cite{xia2017aid}
with the multi-label annotations of \cite{hua2020aidml} contains 3000 images
of $600\times600$ pixels collected from multiple sensors at GSDs ranging
from about $0.5$~m to $8$~m. The DFC15 set originates from the 2015 IEEE GRSS
Data Fusion Contest over Zeebrugge \cite{dfc15} and comprises 3342 tiles of
$600\times600$ pixels at $0.05$~m GSD with the multi-label annotations of
\cite{hua2019dfc15mlc}. Vocabularies are harmonized by mapping
\emph{impervious} (DFC15) to \emph{pavement} and \emph{boat} to \emph{ship};
labels outside the shared vocabulary are discarded. Table~\ref{tbl:datasets}
summarizes the resulting per-domain statistics. The three domains differ
markedly in GSD, sensor characteristics, geographic region, and label
priors, e.g.~\emph{tree} is frequent in UCM and AID but rare in
DFC15, which makes the benchmark a demanding test of multi-label
generalization. Fig.~\ref{fig:domain_samples} presents the shared classes from all domains in order to better see the variations across domains.

\begin{table}[t]
\caption{Benchmark statistics over the six shared labels. Counts are numbers
of positive images per label; GSD denotes ground sampling distance in
meters.}\label{tbl:datasets}
\begin{center}
\footnotesize
\setlength{\tabcolsep}{4.5pt}
\begin{tabular}{lrrr}
\toprule
 & UCM \cite{yang2010ucm,chaudhuri2018ucmml}
 & AID \cite{xia2017aid,hua2020aidml}
 & DFC15 \cite{dfc15,hua2019dfc15mlc} \\
\midrule
Images            & 2100            & 3000              & 3342 \\
Tile size (px)    & $256\times256$  & $600\times600$    & $600\times600$ \\
GSD (m)           & $\approx0.3$    & $\approx0.5$--$8$ & $0.05$ \\
Labels per image  & 2.00            & 3.35              & 1.90 \\
\midrule
building          & 691             & 2161              & 1001 \\
car               & 886             & 2026              & 705  \\
tree              & 1009            & 2406              & 258  \\
water             & 203             & 852               & 998  \\
pavement          & 1300            & 2328              & 3133 \\
ship              & 102             & 284               & 270  \\
\bottomrule
\end{tabular}
\end{center}
\end{table}

Three leave-one-domain-out splits are defined: Split~1 trains on
\{AID, DFC15\} and tests on UCM; Split~2 trains on \{UCM, DFC15\} and
tests on AID; Split~3 trains on \{UCM, AID\} and tests on DFC15. Each
domain is partitioned into training, validation, and test subsets
($80/10/10\%$) with a seed-dependent split shared by all methods; source
training and validation subsets are used for learning,
and the target domain is evaluated on its held-out test subset only.

\subsection{Evaluation metrics}\label{sec:metrics}

Performance is measured with threshold-free and thresholded multi-label
metrics \cite{zhang2014review}: per-label average precision and its mean
(mAP), the macro-averaged per-class F1 score (CF1), and the micro-averaged
overall F1 score (OF1). For CF1 and OF1, a per-label decision threshold is
selected on the source validation sets by maximizing the per-label F1 and is
then kept fixed for target evaluation, so no target information enters
threshold selection. All main results are reported as the mean and standard
deviation over five seeds, which control data splits, initialization, and
augmentation randomness.

\subsection{Compared methods}\label{sec:baselines}

The six LD variants are compared against empirical risk minimization (ERM)
with binary cross-entropy; the three corresponding global-statistics
methods: MixStyle \cite{zhou2021mixstyle}, EFDMix \cite{zhang2022efdm},
and CSU \cite{zhang2024csu}, all inserted at the same backbone stages; and
SWAD \cite{cha2021swad} as a representative optimization-based DG method.
An \emph{oracle} trained and evaluated within the target domain (its
training/test partitions) indicates the within-domain ceiling. All methods
share the same backbone, optimizer, schedule, data augmentations, and evaluation
pipeline; they differ only in the components under study.

\subsection{Implementation details}\label{sec:implementation}

All experiments use a ResNet-50 backbone \cite{he2016resnet} pre-trained on
ImageNet \cite{deng2009imagenet}, with inputs resized to $224\times224$
pixels and augmented with horizontal/vertical flips and color jittering.
Models are trained with AdamW \cite{loshchilov2019adamw} (backbone learning
rate $1.16\times10^{-4}$, classifier-head learning rate $9.39\times10^{-4}$,
weight decay $1.59\times10^{-5}$), a cosine schedule decaying to $10^{-6}$,
gradient-norm clipping at $1.0$, and a batch size of $32$ drawn by the
domain-balanced label-matched sampler of Section~\ref{sec:pairing}. Training
runs for at most 100 epochs with early stopping (patience 15) on the
source-validation mAP. The supplementary training cost of the LD modules is
analyzed in Section~\ref{sec:complexity}.

\subsection{Hyperparameter selection}\label{sec:hpo}

Hyperparameters are chosen with a leave-one-source-out (LOSO) proxy objective in the spirit
of \cite{gulrajani2021domainbed}: each source domain is held out in turn,
the model is trained on the remaining sources, and the average held-out
validation mAP constitutes the search objective. With only 2 source domains available per split, LOSO was constructed by pairing the remaining source domain with a heavily augmented version of itself as a pseudo domain. Searches are conducted with
the tree-structured Parzen estimator of Optuna \cite{akiba2019optuna}
(multivariate mode, median pruning): 40 trials for the shared optimizer
parameters, reused by all methods, and 20 trials per method for the
method-specific parameters listed in Table~\ref{tbl:search}. For some LD
variants, the LOSO-selected values were further adjusted to push for more aggressive operations.
Table~\ref{tbl:selected} lists the selected values of the headline variant.

\begin{table}[t]
\caption{Hyperparameter search spaces. ``Log'' denotes log-uniform sampling;
$\mathcal{U}$ denotes uniform sampling.}\label{tbl:search}
\begin{center}
\footnotesize
\setlength{\tabcolsep}{3.5pt}
\begin{tabular}{lll}
\toprule
Parameter & Range & Applies to \\
\midrule
backbone learning rate & $[10^{-6},5\times10^{-4}]$, log & All \\
head learning rate     & $[10^{-5},5\times10^{-2}]$, log & All \\
weight decay           & $[10^{-6},5\times10^{-1}]$, log & All \\
\midrule
$\alpha$ (Beta Dist.)        & $\mathcal{U}[0.1,5.0]$ & (LD-)MixStyle  \\
& & \& (LD-)EFDMix \\
gate probability $p$   & $\mathcal{U}[0.2,0.8]$ & All augmentations \\
insertion stages $\mathcal{K}$ & subsets of $\{1,\dots,4\}$ & LD variants \\
\midrule
warm-up $W$ (epochs)   & $\{0,\dots,10\}$ & All LD variants \\
diversity weight $w_{\mathrm{div}}$ & $[0.01,10]$, log & LLAM variants \\
temperature $\tau$     & $[0.1,100]$, log & LLAM variants \\
bank size $M$          & $\{64,128,256,512\}$ & LD-MixStyle \\
top-$K$ ratio $\rho$   & $\mathcal{U}[0.01,1.0]$ & LD-EFDMix \\
strength $\beta$       & $\mathcal{U}[0.1,1.0]$ & (LD-)CSU \\
refresh $R$ (epochs)   & $\{1,\dots,15\}$ & GC variants \\
\bottomrule
\end{tabular}
\end{center}
\end{table}

\begin{table}[t]
\caption{Selected hyperparameters of LD-EFDMix-GC per split
(S1--S3).}\label{tbl:selected}
\begin{center}
\footnotesize
\setlength{\tabcolsep}{5pt}
\begin{tabular}{lccc}
\toprule
Parameter & S1 & S2 & S3 \\
\midrule
insertion stages $\mathcal{K}$ & $\{2\}$ & $\{1\}$ & $\{1\}$ \\
$\alpha$ (Beta)        & 0.45 & 5.0 & 5.0 \\
gate probability $p$   & 0.20 & 0.80 & 0.80 \\
top-$K$ ratio $\rho$   & 0.34 & 0.60 & 0.60 \\
refresh $R$ (epochs)   & 9    & 13  & 13  \\
warm-up $W$ (epochs)   & 9    & 5   & 5   \\
\bottomrule
\end{tabular}
\end{center}
\end{table}

% ============================================================================
\section{Results and discussion}\label{sec:results}
% ============================================================================

\subsection{Comparison with global-statistics augmentation}\label{sec:main}

Tables~\ref{tbl:proto1}--\ref{tbl:proto3} report per-label average precision,
mAP, CF1, and OF1 for the three splits, and Table~\ref{tbl:summary}
averages the mAP results. Several observations can be discussed.

\begin{table*}[t]
\caption{Split 1 (sources: AID, DFC15; target: UCM). Per-label average
precision and aggregate metrics in $\%$ (mean$\pm$std over five seeds;
per-label columns show means). Best value per column among the domain
generalization methods in bold; the within-domain oracle is reported for
reference only.}\label{tbl:proto1}
\begin{center}
\footnotesize
\setlength{\tabcolsep}{4.5pt}
\begin{tabular*}{\textwidth}{@{\extracolsep{\fill}}lrrrrrrccc@{}}
\toprule
Method & building & car & tree & water & pavement & ship & mAP & CF1 & OF1 \\
\midrule
ERM (BCE) & 58.2 & 71.9 & 83.4 & 45.9 & 84.4 & 87.9 & 72.0$\pm$1.2 & 69.5$\pm$1.5 & 73.1$\pm$1.4 \\
\midrule
MixStyle (ICLR 2021) \cite{zhou2021mixstyle} & 68.7 & 77.7 & 81.8 & 62.9 & \textbf{87.5} & 93.5 & 78.7$\pm$1.2 & 73.0$\pm$3.0 & 74.5$\pm$2.2 \\
EFDMix (CVPR 2022) \cite{zhang2022efdm} & 68.1 & 73.8 & 85.0 & 63.6 & 85.3 & 93.4 & 78.2$\pm$4.3 & 73.0$\pm$3.4 & 74.6$\pm$1.7 \\
CSU (WACV 2024) \cite{zhang2024csu} & 70.6 & 76.6 & 83.2 & 52.2 & 85.7 & 88.3 & 76.1$\pm$1.5 & 71.2$\pm$1.0 & 73.7$\pm$1.8 \\
SWAD (NeurIPS 2021) \cite{cha2021swad} & 56.3 & 71.3 & 85.0 & 47.5 & 84.3 & 87.4 & 72.0$\pm$1.8 & 69.3$\pm$2.2 & 72.4$\pm$1.9 \\
\midrule
LD-MixStyle & 67.8 & 72.4 & 85.4 & \textbf{65.3} & 84.0 & 90.4 & 77.6$\pm$3.6 & 73.5$\pm$2.6 & 74.6$\pm$1.9 \\
LD-EFDMix & 69.7 & 76.5 & \textbf{86.9} & 61.8 & 83.8 & 92.4 & 78.5$\pm$2.3 & 73.8$\pm$1.6 & \textbf{75.0$\pm$1.1} \\
LD-CSU & 67.9 & 78.4 & 85.2 & 55.2 & 85.2 & 95.2 & 77.9$\pm$1.8 & 73.1$\pm$1.2 & 74.9$\pm$1.9 \\
LD-MixStyle-GC & 68.0 & 74.3 & 85.6 & 61.2 & 84.4 & 96.2 & 78.3$\pm$2.9 & 73.7$\pm$2.0 & \textbf{75.0$\pm$2.5} \\
LD-EFDMix-GC & \textbf{71.8} & \textbf{79.0} & 84.9 & 61.0 & 83.5 & \textbf{96.6} & \textbf{79.5$\pm$2.4} & \textbf{74.1$\pm$2.4} & 74.9$\pm$2.2 \\
LD-CSU-GC & 69.0 & 77.3 & 85.3 & 49.4 & 85.1 & 91.9 & 76.3$\pm$3.3 & 72.1$\pm$1.9 & 74.0$\pm$0.8 \\
\midrule
Oracle (within-domain) & 93.8 & 95.6 & 95.7 & 98.5 & 96.5 & 99.8 & 96.7$\pm$0.3 & 90.1$\pm$0.6 & 88.6$\pm$0.5 \\
\bottomrule
\end{tabular*}
\end{center}
\end{table*}

\begin{table*}[t]
\caption{Split 2 (sources: UCM, DFC15; target: AID). Notation as in
Table~\ref{tbl:proto1}.}\label{tbl:proto2}
\begin{center}
\footnotesize
\setlength{\tabcolsep}{4.5pt}
\begin{tabular*}{\textwidth}{@{\extracolsep{\fill}}lrrrrrrccc@{}}
\toprule
Method & building & car & tree & water & pavement & ship & mAP & CF1 & OF1 \\
\midrule
ERM (BCE) & 91.6 & 92.2 & 92.6 & 62.5 & 93.7 & 32.4 & 77.5$\pm$2.1 & 55.7$\pm$4.1 & 67.2$\pm$3.8 \\
\midrule
MixStyle \cite{zhou2021mixstyle} & 93.0 & 92.4 & \textbf{93.6} & 61.9 & 92.3 & 27.4 & 76.8$\pm$1.6 & 57.3$\pm$1.7 & \textbf{70.8$\pm$1.1} \\
EFDMix \cite{zhang2022efdm} & 93.1 & 89.1 & 93.1 & 58.5 & 91.1 & 36.6 & 76.9$\pm$1.8 & 56.8$\pm$6.1 & 69.1$\pm$3.8 \\
CSU \cite{zhang2024csu} & 93.1 & 92.7 & 92.8 & 63.1 & \textbf{95.6} & 32.1 & 78.2$\pm$1.0 & \textbf{60.2$\pm$2.8} & 70.1$\pm$2.9 \\
SWAD \cite{cha2021swad} & 91.5 & 92.6 & 92.8 & \textbf{63.5} & 93.7 & 29.9 & 77.3$\pm$2.1 & 54.6$\pm$3.8 & 66.7$\pm$3.9 \\
\midrule
LD-MixStyle & 91.3 & 91.7 & 91.5 & 58.3 & 93.5 & \textbf{43.4} & 78.3$\pm$1.6 & 57.3$\pm$5.5 & 66.4$\pm$3.8 \\
LD-EFDMix & 94.5 & 92.5 & 92.5 & 61.7 & 92.6 & 40.8 & \textbf{79.1$\pm$1.6} & 56.0$\pm$2.0 & 69.1$\pm$1.5 \\
LD-CSU & 92.8 & 92.5 & 91.4 & 59.6 & 94.7 & 39.3 & 78.4$\pm$1.6 & 54.5$\pm$4.5 & 66.4$\pm$3.6 \\
LD-MixStyle-GC & \textbf{95.2} & 92.0 & 91.7 & 54.5 & 94.6 & 36.0 & 77.3$\pm$1.6 & 55.6$\pm$3.7 & 70.1$\pm$1.7 \\
LD-EFDMix-GC & 93.6 & 92.9 & 93.2 & 60.9 & 90.8 & 36.0 & 77.9$\pm$2.2 & 58.0$\pm$4.1 & 69.3$\pm$5.4 \\
LD-CSU-GC & 93.9 & \textbf{93.1} & 92.9 & 58.6 & 93.3 & 38.3 & 78.4$\pm$1.3 & 55.1$\pm$2.2 & 67.5$\pm$2.8 \\
\midrule
Oracle (within-domain) & 99.3 & 98.3 & 98.9 & 82.5 & 99.5 & 76.5 & 92.5$\pm$0.4 & 86.5$\pm$1.5 & 92.8$\pm$0.8 \\
\bottomrule
\end{tabular*}
\end{center}
\end{table*}

\begin{table*}[t]
\caption{Split 3 (sources: UCM, AID; target: DFC15). Notation as in
Table~\ref{tbl:proto1}.}\label{tbl:proto3}
\begin{center}
\footnotesize
\setlength{\tabcolsep}{4.5pt}
\begin{tabular*}{\textwidth}{@{\extracolsep{\fill}}lrrrrrrccc@{}}
\toprule
Method & building & car & tree & water & pavement & ship & mAP & CF1 & OF1 \\
\midrule
ERM (BCE) & 43.1 & 42.3 & 13.5 & 62.1 & 96.1 & 43.7 & 50.1$\pm$6.2 & 46.2$\pm$4.2 & 60.4$\pm$2.3 \\
\midrule
MixStyle \cite{zhou2021mixstyle} & 46.4 & 36.3 & 25.4 & 58.5 & 97.8 & 54.7 & 53.2$\pm$4.0 & 47.0$\pm$3.8 & 61.9$\pm$1.8 \\
EFDMix \cite{zhang2022efdm} & 48.5 & 42.8 & 17.3 & 55.9 & 98.0 & 48.8 & 51.9$\pm$3.8 & 46.8$\pm$2.0 & 63.4$\pm$2.1 \\
CSU \cite{zhang2024csu} & 51.0 & 41.3 & 24.3 & 63.5 & 98.2 & \textbf{59.7} & 56.3$\pm$2.6 & 49.4$\pm$1.9 & 64.2$\pm$1.9 \\
SWAD \cite{cha2021swad} & 42.1 & 40.0 & 11.0 & 60.6 & 96.9 & 44.6 & 49.2$\pm$5.6 & 45.9$\pm$5.6 & 60.6$\pm$2.7 \\
\midrule
LD-MixStyle & \textbf{60.0} & 48.6 & 23.4 & 61.4 & 97.4 & 52.7 & 57.3$\pm$3.8 & 49.4$\pm$3.2 & 64.0$\pm$1.7 \\
LD-EFDMix & 55.9 & 44.1 & 21.9 & 65.2 & 98.1 & 54.7 & 56.7$\pm$3.1 & 47.9$\pm$3.3 & 64.6$\pm$2.2 \\
LD-CSU & 45.3 & 47.8 & \textbf{30.7} & \textbf{69.4} & 98.2 & 55.4 & \textbf{57.8$\pm$2.5} & \textbf{50.9$\pm$2.3} & \textbf{65.9$\pm$2.1} \\
LD-MixStyle-GC & 51.8 & 42.8 & 17.1 & 61.3 & 98.3 & 54.0 & 54.2$\pm$4.3 & 49.5$\pm$3.1 & 65.6$\pm$2.5 \\
LD-EFDMix-GC & 51.7 & \textbf{52.0} & 21.9 & 61.8 & \textbf{98.4} & 57.8 & 57.3$\pm$5.1 & 50.8$\pm$3.2 & 65.5$\pm$1.0 \\
LD-CSU-GC & 54.2 & 43.4 & 20.4 & 66.2 & 97.8 & 57.7 & 56.6$\pm$2.6 & 48.3$\pm$2.7 & 62.7$\pm$2.9 \\
\midrule
Oracle (within-domain) & 98.6 & 97.7 & 92.8 & 99.1 & 100.0 & 94.8 & 97.2$\pm$1.1 & 91.9$\pm$2.0 & 95.6$\pm$0.6 \\
\bottomrule
\end{tabular*}
\end{center}
\end{table*}

\begin{table}[t]
\caption{Target-domain mAP ($\%$) average over splits, and the difference of that average with respect
to ERM. Best value per column among the domain generalization methods in
bold.}\label{tbl:summary}
\begin{center}
\footnotesize
\setlength{\tabcolsep}{6pt}
\begin{tabular}{lcc}
\toprule
Method & Average & $\Delta$ vs.\ ERM \\
\midrule
ERM (BCE) & 66.5 & -- \\
\midrule
MixStyle \cite{zhou2021mixstyle} & 69.5 & +3.0 \\
EFDMix \cite{zhang2022efdm} & 69.0 & +2.5 \\
CSU \cite{zhang2024csu} & 70.2 & +3.7 \\
SWAD \cite{cha2021swad} & 66.2 & $-$0.4 \\
\midrule
LD-MixStyle & 71.0 & +4.5 \\
LD-EFDMix & 71.4 & +4.9 \\
LD-CSU & 71.3 & +4.8 \\
LD-MixStyle-GC & 69.9 & +3.4 \\
LD-EFDMix-GC & \textbf{71.5} & \textbf{+5.0} \\
LD-CSU-GC & 70.4 & +3.9 \\
\midrule
Oracle (within-domain) & 95.4 & +28.9 \\
\bottomrule
\end{tabular}
\end{center}
\end{table}

First, statistics-based augmentation is clearly beneficial in this setting:
the global methods improve the average mAP over ERM by $2.5$--$3.7$ points,
whereas SWAD performs on par with ERM ($66.2$ vs.~$66.5$), suggesting that
flat-minima optimization alone does not address the dominant appearance shift
between these aerial domains.

Second, label decoupling improves every operator it is applied to. With
learned attention, LD-MixStyle, LD-EFDMix, and LD-CSU exceed their global
counterparts by $+1.5$, $+2.4$, and $+1.1$ mAP points on average,
respectively; with the Grad-CAM bank, LD-EFDMix-GC gains $+2.5$ over EFDMix
and attains the best overall average of $71.5\%$ ($+5.0$ over ERM, $+1.3$
over the strongest global baseline, CSU). The gains are most pronounced on
Split~3, the hardest transfer (target DFC15 at $0.05$~m GSD): ERM drops to
$50.1\%$, while LD-CSU reaches $57.8\%$ ($+7.7$) and LD-EFDMix-GC and
LD-MixStyle reach $57.3\%$. The advantage of decoupling also appears at the
label level: on Split~2, ERM and the global methods nearly collapse on the
minority class \emph{ship} ($27$--$37\%$ average precision), whereas the
LLAM-based LD variants recover $39$--$43\%$, consistent with the premise that
restricting style exchange to label-matched regions protects rare classes
from being overwritten by dominant ones.

Third, the benefit of the cached attention source is operator-dependent. For
EFDMix, whose sort matching directly consumes the attention-selected
locations, the Grad-CAM bank is the strongest configuration on
Split~1 ($79.5\%$) and on average; for MixStyle and CSU, the cached maps
help less than the jointly trained LLAM ($+0.4$ and $+0.2$ average gain over
the global counterparts, against $+1.5$ and $+1.1$ with LLAM). A plausible
explanation is that moment pooling integrates over the whole map, so it
benefits more from attention that is optimized end-to-end together with the
mixing, whereas rank matching mainly requires that the \emph{support} of the
map be correct, which the Grad-CAM bank provides cheaply.

Fourth, Split~2 exhibits the smallest margins for all augmentation
methods ($76.8$--$79.1\%$ around an ERM of $77.5\%$). AID is itself a
multi-sensor, multi-resolution collection \cite{xia2017aid}, so its style
distribution overlaps the sources more than in the other splits, leaving
less room for style-based augmentation. Finally, the gap to the
within-domain oracle remains large ($13$--$40$ mAP points depending on the
split), indicating that style augmentation alone, whether global or
label-decoupled, closes only part of the domain gap; the limitation analysis
in Section~\ref{sec:limitations} returns to this point. This is expected:
domain shift between these datasets is not purely a matter of style but also
reflects differences in ground sampling distance, sensor and band
characteristics, and class co-occurrence statistics (Table~\ref{tbl:datasets}),
which channel-statistics operators are not designed to correct. A style-centric
method is nonetheless the appropriate tool for the appearance component of the
gap that the AdaIN family does model, and the manifold analysis of
Section~\ref{sec:manifold} indicates that the per-label style component is the
part that aligns across domains; the residual, structural component is what the
distance to the oracle reflects.

\subsection{Ablation study}\label{sec:ablation}

Table~\ref{tbl:ablation} removes the components of LD-EFDMix-GC one at a time
on Split~1, and additionally evaluates LD-MixStyle without its style bank.
Replacing the per-label attention by a uniform map (which collapses the
decomposition toward global statistics) costs $2.6$ mAP points, and freezing
the Grad-CAM bank after its first construction costs $3.3$ points, indicating that spatially correct \emph{and} up-to-date
localization is the core of the method. Disabling the top-$K$ restriction
($\rho=1$) costs $1.6$ points, supporting the view that sort matching should
be confined to label-relevant locations. The remaining components contribute
smaller but consistent amounts: per-label coefficients ($-0.8$),
label-matched pairing ($-0.6$), and the warm-up ($-0.5$). The style bank of
LD-MixStyle yields a modest $+0.3$ over batch-only pairing, suggesting that
the domain-balanced label-matched sampler already supplies sufficient
cross-domain partners in this benchmark.

\begin{table*}[t]
\caption{Component ablations on Split 1 ($\%$, mean$\pm$std over five
seeds). $\Delta$mAP is the change with respect to the corresponding full
model.}\label{tbl:ablation}
\begin{center}
\footnotesize
\setlength{\tabcolsep}{6pt}
\begin{tabular*}{\textwidth}{@{\extracolsep{\fill}}lcccc@{}}
\toprule
Variant & mAP & CF1 & OF1 & $\Delta$mAP \\
\midrule
LD-EFDMix-GC (full) & \textbf{79.5$\pm$2.4} & \textbf{74.1$\pm$2.4} & 74.9$\pm$2.2 & -- \\
w/o attention (uniform map) & 76.8$\pm$3.0 & 72.5$\pm$0.8 & 74.2$\pm$2.3 & $-$2.6 \\
w/o map refresh & 76.2$\pm$3.5 & 72.6$\pm$1.2 & 74.4$\pm$1.3 & \textbf{$-$3.3} \\
w/o top-$K$ selection ($\rho=1$) & 77.8$\pm$2.9 & 71.8$\pm$1.1 & 73.7$\pm$1.3 & $-$1.6 \\
w/o per-label $\lambda$ (shared $\lambda$) & 78.6$\pm$2.4 & 73.7$\pm$2.5 & \textbf{75.6$\pm$2.4} & $-$0.8 \\
w/o label-matched pairing & 78.8$\pm$1.4 & 74.1$\pm$2.2 & 75.1$\pm$2.4 & $-$0.6 \\
w/o warm-up & 79.0$\pm$2.0 & 73.2$\pm$1.7 & 75.5$\pm$1.3 & $-$0.5 \\
\midrule
LD-MixStyle (full) & \textbf{77.6$\pm$3.6} & \textbf{73.5$\pm$2.6} & \textbf{74.6$\pm$1.9} & -- \\
LD-MixStyle w/o style bank & 77.3$\pm$1.5 & 72.0$\pm$2.1 & 74.4$\pm$2.1 & $-$0.3 \\
\bottomrule
\end{tabular*}
\end{center}
\end{table*}

\subsection{Hyperparameter sensitivity}\label{sec:sensitivity}

\begin{figure*}[t]
\begin{center}
\subfloat[Insertion point]{\label{fig:sens-a}\includegraphics[width=0.30\textwidth]{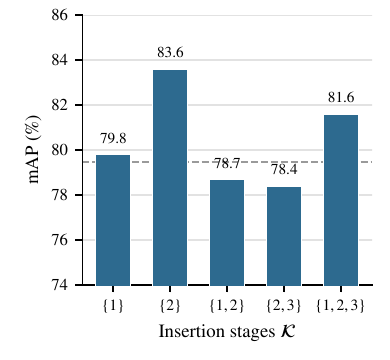}}\hfil
  \subfloat[Top-$K$ ratio]{\label{fig:sens-b}\includegraphics[width=0.30\textwidth]{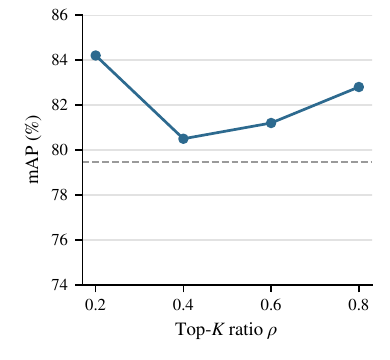}}\hfil
  \subfloat[Map refresh interval]{\label{fig:sens-c}\includegraphics[width=0.30\textwidth]{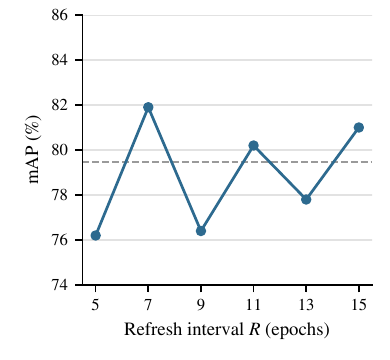}}\\[2pt]
  \subfloat[LLAM temperature]{\label{fig:sens-d}\includegraphics[width=0.45\textwidth]{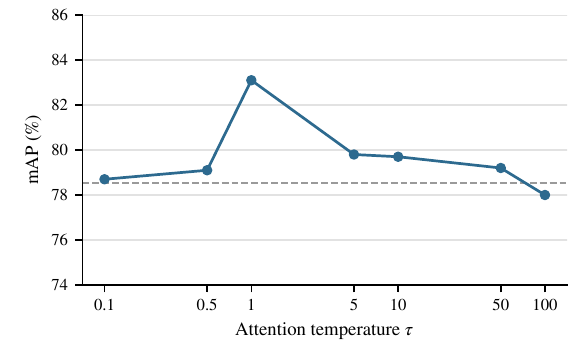}}\hfil
  \subfloat[Style-bank size]{\label{fig:sens-e}\includegraphics[width=0.45\textwidth]{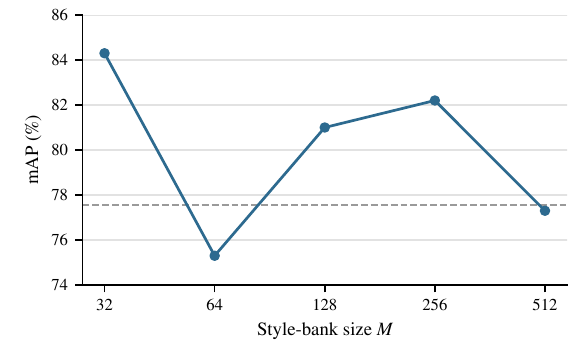}}
\end{center}
\caption{Hyperparameter sensitivity on Split 1 (target: UCM); each point
is a single training run at a fixed seed shared by all sensitivity runs and different from any training run seed. (a)--(c) LD-EFDMix-GC: insertion
stages, top-$K$ ratio $\rho$, and Grad-CAM refresh interval $R$; (d)
LD-EFDMix: LLAM temperature $\tau$; (e) LD-MixStyle: style-bank size $M$. Dashed lines show the five-seed
mean of the training configuration. The refresh and top-$K$ sweeps use a one-epoch
warm-up so that the studied mechanism governs most of the run.}
\label{fig:sensitivity}
\end{figure*}

Fig.~\ref{fig:sensitivity} examines the main hyperparameters on Split~1
with single training runs at a fixed seed different from any of the main 5 seeds. 
The insertion point is the most
influential choice: a single module after stage~2 performs best ($83.6\%$ in
this run), early-plus-deep combinations underperform, and the all-stage
configuration lies in between; consistent with the common observation that
style information concentrates in early-to-middle layers
\cite{zhou2021mixstyle,li2022dsu}. The top-$K$ ratio favors selective
matching: the smallest tested value ($\rho=0.2$) performs best, and quality
degrades toward $\rho=1$, in line with the corresponding ablation. The
refresh interval shows no consistent trend over $R\in[5,15]$; the spread
($76$--$82\%$) is comparable to the seed-to-seed variability of the full
model ($\pm2.4$), suggesting that any sufficiently frequent refresh is
adequate, while never refreshing is harmful (Table~\ref{tbl:ablation}). The
LLAM temperature exhibits a broad plateau for $\tau\in[1,50]$ with a mild
decline at $\tau=100$, where overly sharp maps may discard context. The bank
size likewise shows no consistent trend within run-to-run variability,
matching its small ablation effect. Overall, apart from the insertion point,
the method appears robust to its hyperparameters around the selected values.

\subsection{Style-manifold analysis}\label{sec:manifold}

\begin{figure*}[t]
\begin{center}
\includegraphics[width=\textwidth]{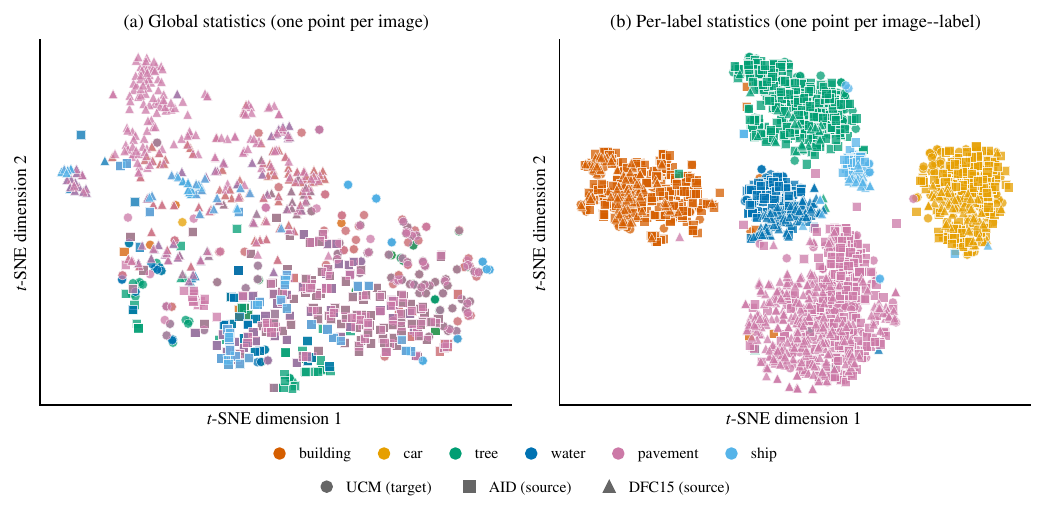}
\end{center}
\caption{$t$-SNE \cite{vandermaaten2008tsne} embeddings of style statistics
for an LD-EFDMix model on Split 1 (test partitions of all domains).
(a)~Global per-image statistics $(\boldsymbol{\mu},\boldsymbol{\sigma})$,
plotted once per (image, present label) pair for color comparability.
(b)~Per-label statistics
$(\boldsymbol{\mu}_{\ell},\boldsymbol{\sigma}_{\ell})$ from the proposed
decomposition. Colors denote labels; marker shapes denote domains.}
\label{fig:manifold}
\end{figure*}

Fig.~\ref{fig:manifold} visualizes, with $t$-SNE \cite{vandermaaten2008tsne},
the style statistics that the two augmentation philosophies operate on, using
an LD-EFDMix model from Split~1 and the test partitions of all three
domains. When global per-image statistics are embedded
(Fig.~\ref{fig:manifold}a), points organize primarily by domain, and images
carrying different labels from the same domain remain close together: the
quantity that global methods mix is dominated by sensor- and scene-level
appearance. The per-label statistics extracted by the proposed decomposition
(Fig.~\ref{fig:manifold}b) instead form label-coherent groupings within which
the three domain markers interleave, and target-domain points fall inside or
near the groups spanned by the sources. This pattern supports the central
premise of the method: per-label pooling exposes label-specific style
components that are comparable across domains, which is precisely the
structure that label-matched mixing exploits.

\subsection{Complexity analysis}\label{sec:complexity}

Table~\ref{tbl:complexity} compares model complexity. The LLAM variants add
$8.3\times10^{4}$ parameters ($+0.35\%$) for the configuration with modules
after stages~1 and~2; the GC variants add none. Because all modules are
identities at inference, the multiply--accumulate count is unchanged
($4.087$~G at $224\times224$ pixels) and the measured evaluation throughput
is statistically indistinguishable across methods ($1495$--$1505$ images/s
at batch size 32 on the GPU used in this study). Training-time overheads are
confined to the augmentation pathway: LLAM adds two $1\times1$ convolutions
per inserted stage and the diversity loss, while the GC variants add
approximately one extra forward--backward sweep over the training set every
$R$ epochs to rebuild the attention bank. These properties make the
framework a near drop-in replacement for its global counterparts.

\begin{table}[t]
\caption{Complexity at $224\times224$-pixel inputs. Throughput is measured
in evaluation mode at batch size 32 on a single GPU; MACs denote
multiply--accumulate operations.}\label{tbl:complexity}
\begin{center}
\footnotesize
\setlength{\tabcolsep}{3pt}
\begin{tabular}{lcccc}
\toprule
Method & Params (M) & $\Delta$ & MACs (G) & img/s \\
\midrule
ERM / global    & 23.52 & --        & 4.087 & 1495--1498 \\
LD (LLAM)       & 23.60 & $+0.35\%$ & 4.087 & 1501--1505 \\
LD (Grad-CAM)   & 23.52 & --        & 4.087 & 1498--1500 \\
\bottomrule
\end{tabular}
\end{center}
\end{table}

\subsection{Limitations}\label{sec:limitations}

Several limitations should be noted. First, the gap to the within-domain
oracle remains substantial ($13$--$40$ mAP points), and minority classes with
strong cross-domain appearance variance (\emph{ship} on AID and \emph{tree}
on DFC15) remain difficult for every method evaluated: augmenting feature
statistics cannot synthesize object scales or contexts absent from the
sources. Second, gains on Split~2 are small, indicating that the benefit
of style augmentation shrinks when the target's style distribution already
overlaps the sources. Third, the benchmark covers six shared labels and
three domains; transfer of the conclusions to larger vocabularies and other
sensor families remains to be verified due to lack of similar annotated multi-label multi-domain data in the literature. Fourth, the sensitivity study relies
on single runs per configuration, so only coarse trends should be read from
it. Fifth, all methods are compared under a common ResNet-50 backbone with
matched training, but the study does not include large vision-language or
foundation-model baselines; in general vision, multi-label DG is increasingly
driven by such models (e.g.~CLIPood \cite{shu2023clipood} and Mixup-CLIPood
\cite{qiao2024mixupclipood}), and geospatial vision foundation models have
begun to enter RS DG (e.g.~CrossEarth \cite{gong2025crossearth}). A controlled
comparison at that capacity is an important direction left to future work, as it
entails a markedly heavier and different training and fine-tuning regime;
label-decoupled statistics are, in any case, an orthogonal mechanism that could
be inserted into such backbones rather than a competitor to them.
Finally, the method assumes that image-level labels suffice to localize
label-specific style; in scenes where a class occupies very few pixels, the
attention, learned or cached, may be too coarse.

% ============================================================================
\section{Conclusion}\label{sec:conclusion}
% ============================================================================

This paper introduced a label-decoupled style augmentation framework for
domain generalization in multi-label remote sensing image classification.
The framework replaces the global feature statistics used by MixStyle,
EFDMix, and CSU with attention-derived per-label statistics, mixes them only
between cross-domain samples sharing the corresponding label with independent
per-label coefficients, and recomposes the feature map through
attention-weighted normalization; the attention is provided either by a
lightweight learned module or by a periodically refreshed Grad-CAM bank. On a
leave-one-domain-out benchmark assembled from the multi-label UCM, AID, and
DFC15 datasets, the proposed variants improved upon their global counterparts
in all configurations, with the best variant reaching $71.5\%$ average mAP
($+5.0$ over ERM, and up to $+7.7$ on the hardest transfer), at a cost of at
most $0.35\%$ additional parameters and, as for other statistics-based
augmenters, no change at inference. Ablations and
a manifold analysis indicate that spatially correct, up-to-date per-label
localization is the decisive component, and that per-label statistics are
considerably better aligned across domains than their global counterparts.

The results suggest that label decoupling is a generic and inexpensive
upgrade for statistics-based augmentation whenever images contain several
co-occurring classes. Future work could examine larger shared vocabularies
and additional sensor families, the behavior of the decomposition on
transformer backbones, the combination of label-decoupled augmentation with
optimization-based methods such as weight averaging, and the integration of
label-relation priors into the pairing and bank mechanisms; extending the
attention sources to handle classes that occupy very few pixels is a further
direction.% 

% ============================================================================
\bibliographystyle{unsrt}
\bibliography{refs}
\end{document}